\documentclass[conference,letterpaper,10pt]{ieeeconf}
\usepackage[left=1.91cm,right=1.91cm,top=1.91cm,bottom=1.91cm]{geometry}

\usepackage{graphicx}
\usepackage{amssymb}
\usepackage{amsmath}
\usepackage{cite}
\usepackage{comment}
\usepackage{color}
\usepackage{url}
\usepackage{alltt}
\usepackage{cleveref}
\usepackage{tikz}
\usepackage{pgfplots}
\usepackage{pgf-umlsd}
\usepackage{cprotect}
\usepackage{multirow}
\usepackage{algorithm}
\usepackage[noend]{algpseudocode}
\usepackage{bigfoot}
\usepackage{paralist}
\usepackage{tabularx}
\usepackage[whole]{bxcjkjatype}
\usepackage{ifthen}
\usepackage{threeparttable}
\usepackage[hang,small,bf]{caption}
\usepackage[subrefformat=parens]{subcaption}
\usepackage[super]{nth}
\usepackage{colortbl}
\usepackage{scalefnt}
\usepackage{threeparttable}

\pgfplotsset{compat=1.14}
\IEEEoverridecommandlockouts
\newcommand{\argmax}{\mathop{\rm arg~max}\limits}
\newcommand{\argmin}{\mathop{\rm arg~min}\limits}

\makeatletter
\newcommand{\figcaption}[1]{\def\@captype{figure}\caption{#1}}
\newcommand{\tblcaption}[1]{\def\@captype{table}\caption{#1}}
\makeatother

\title{\LARGE \bf
Goal-Image Conditioned Dynamic Cable Manipulation through Bayesian Inference and Multi-Objective Black-Box Optimization
}

\author{
Kuniyuki Takahashi$^{\dagger}$,
Tadahiro Taniguchi$^{\ddagger}$
\thanks{$^{\dagger}$K.Takahashi is with Preferred Networks, Inc.
        {\tt\footnotesize 
        takahashi@preferred.jp}
        $^{\ddagger}$T.Taniguchi is with Ritsumeikan University, College of Information Science and Engineering.
        {\tt\footnotesize 
        taniguchi@em.ci.ritsumei.ac.jp}
        }
}

\begin{document}

\maketitle
\thispagestyle{empty}

\begin{abstract}
To perform dynamic cable manipulation to realize the configuration specified by a target image, we formulate dynamic cable manipulation as a stochastic forward model.
Then, we propose a method to handle uncertainty by maximizing the expectation, which also considers estimation errors of the trained model.
To avoid issues like multiple local minima and requirement of differentiability by gradient-based methods, we propose using a black-box optimization (BBO) to optimize joint angles to realize a goal image.
Among BBO, we use the Tree-structured Parzen Estimator (TPE), a type of Bayesian optimization.
By incorporating constraints into the TPE, the optimized joint angles are constrained within the range of motion.
Since TPE is population-based, it is better able to detect multiple feasible configurations using the estimated inverse model.
We evaluated image similarity between the target and cable images captured by executing the robot using optimal transport distance.
The results show that the proposed method improves accuracy compared to conventional gradient-based approaches and methods that use deterministic models that do not consider uncertainty.
\footnote{An accompanying video is available at the following link:\\ \url{https://youtu.be/AMDTJRNEbek}}
\end{abstract}
\section{Introduction}
\label{sec:introduction}
Robotic cable manipulation has been used in many fields, such as cable harnessing in factories and sewing operations in the medical field.
However, cables are prone to deform into various shapes during manipulation.

There have been many cable manipulation studies, which can be broadly classified into two research perspectives.
First, we classify cable manipulation into physics-based modeling and learning-based approaches.
Analytical physics-based modeling methods, such as mass-spring systems, position-based dynamics, and finite element methods, are used to model cables~\cite{yin2021modeling}.
These are approximate predetermined models and require accurate cable parameters.
Some methods model an approximate of the cable by tracking the cable from sensor information~\cite{jin2019robust}.
Since these models make a number of assumptions, they sacrifice accuracy and may not be applicable to cables in any environment.
Learning-based approaches can learn directly from data and are useful for flexible objects that are difficult to model.
Therefore, they have the advantage of being applied to arbitrary cables without the need to give a model of the cable.

\begin{figure}[tb]
	\centering
	\includegraphics[width=0.80\columnwidth]{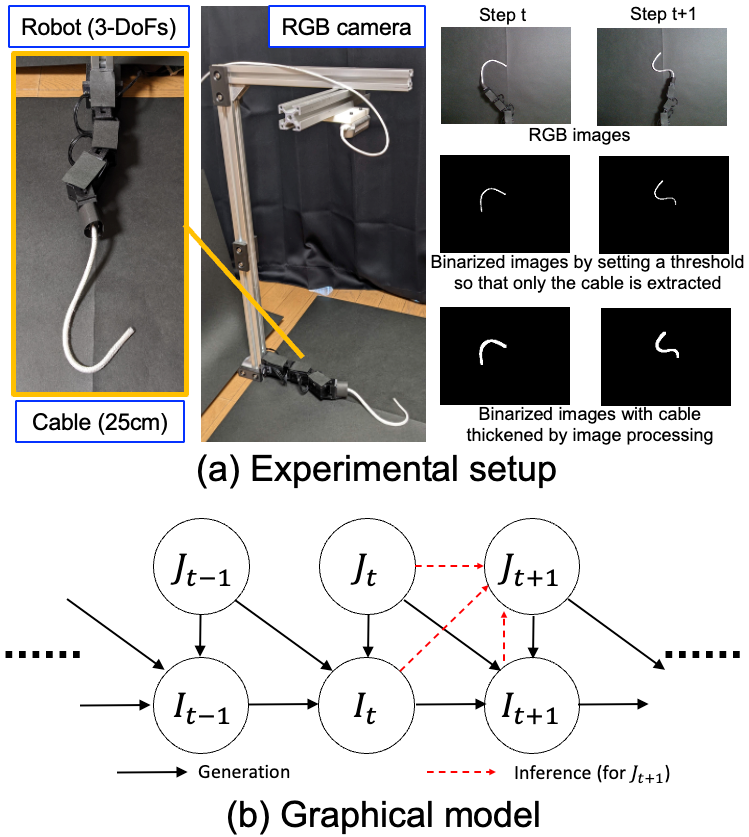}
	\caption{(a) Robot setup used for the experiment of cable manipulation and examples of RGB images and binarized images by setting a threshold so that only the cable is extracted, and binarized images with strings thickened by image processing at step $t$ and step $t+1$, respectively.\newline
            (b) Probabilistic graphical model representing the dependency between the joint angle $J_t$ and the image $I_t$.
    }
	\label{fig:setup}
\end{figure}
\begin{figure*}[t]
	\centering
	\includegraphics[width=1.90\columnwidth]{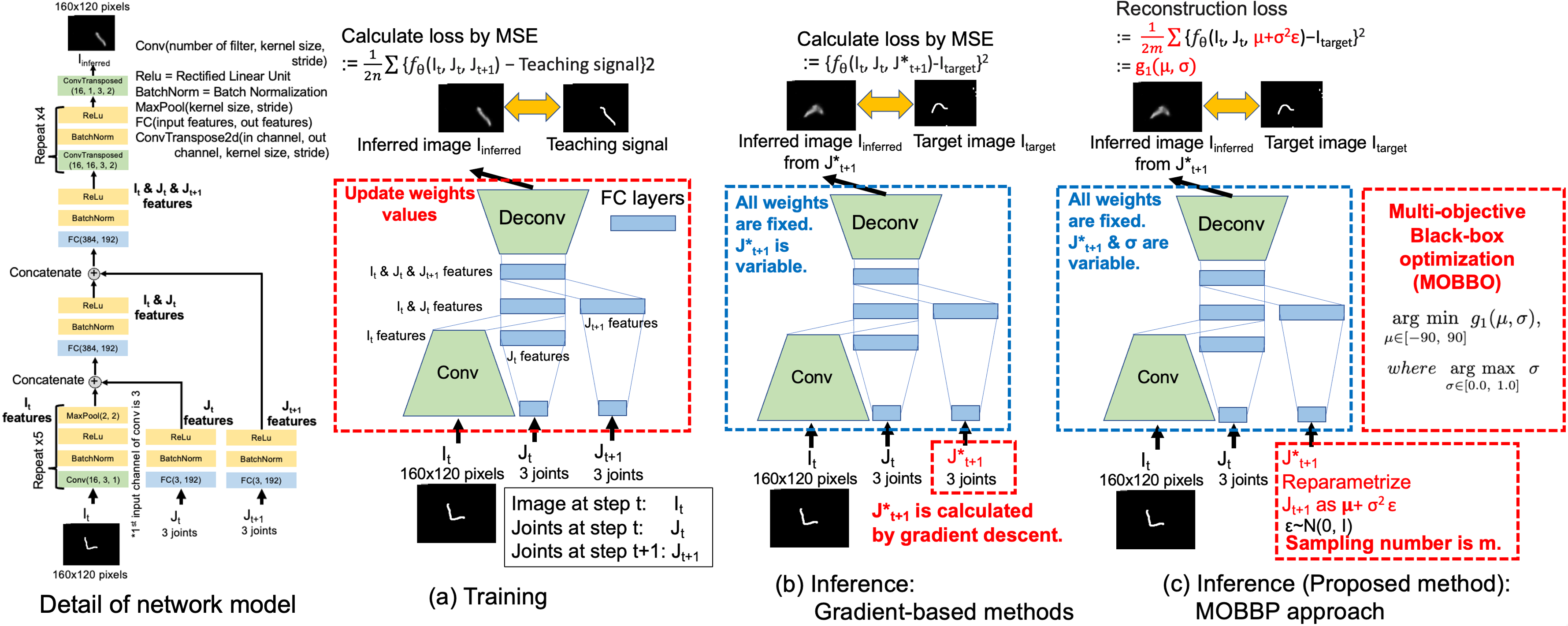}
	\caption{Proposed method and details of the network model}
	\label{fig:method}
    \vspace{-8mm}
\end{figure*}
Next, we classify cable manipulation into their use in static and dynamic tasks.
Cable manipulation is treated as a quasi-static task rather than a static task.
Quasi-static assumes that the motion is slow and there are no inertial effects.
Although the use of deep learning has seen increased adoption in cable manipulation, most tasks assume quasi-static~\cite{nair2017combining, wang2019learning, wu2020learning, yan2020self, sundaresan2020learning, suzuki2021air, yu2022shape, ma2022learning, viswanath2022autonomously}.
For quasi-static tasks, the deterministic approach, which means that once the initial state is determined, all subsequent states are determined, can be handled by modifying the prediction error between each step as necessary.
It is more challenging to modify prediction errors for dynamic motions because of the motion's rapidity.
Furthermore, there are two major sources of uncertainty when dealing with dynamic motion.
The first is that slight differences in friction conditions or delays in the robot's motion in response to commands can cause differences in the state of the cable (aleatory uncertainty).
Second, because flexible objects have infinite states, such that cables change shape in various ways, learning-based approaches require a large amount of data.
However, collecting data that covers every possible outcome is infeasible, leading to possible errors when the model encounters them (epistemic uncertainty).
Some related studies dealing with dynamic motion have used additional learning by readjusting the simulation parameters when errors occur in the model~\cite{lim2022real2sim2real}.
Not many methods deal with dynamic motion yet, but their approaches are deterministic~\cite{kawaharazuka2019dynamic, zhang2021robots, chi2022iterative, lim2022real2sim2real}.
The challenge is that errors between the model and reality can affect the prediction due to uncertainty.
Therefore, the task we address in this research is the dynamic motion of a cable considering uncertainty.

The task is to optimize the target joint angle $J^{\ast}_{t+1}$ of the robotic arm with redundant degrees of freedom attached to a cable to realize the target image $I_{target}$ from the current image $I_{t}$ and joint angle $J_{t}$ (Fig.~\ref{fig:setup}).
The straightforward approach is to directly predict the joint angle $J_{t+1}$ from the current image $I_{t}$, the joint angle $J_{t}$, and the target image $I_{target}$.
This is challenging to train a neural network for because of infinite solutions by redundant degrees of freedom.
Another approach is to prepare a trained neural network that predicts ${I}_{inferred}$ from current image $I_{t}$ and joint angle $J_{t}$ \& $J_{t+1}$, as in the method of \cite{kawaharazuka2019dynamic} (Fig.~\ref{fig:method}~(a)).
Then, when $I_{t}$, $J_{t}$, and $I_{target}$ are given to the neural network, the optimal $J^{\ast}_{t+1}$ is calculated using gradient-based methods with mean squared error (MSE) between $I_{target}$ and $I_{inferred}$ (Fig.~\ref{fig:method}~(b)).

However, there are several challenges in using gradient-based methods.
The neural network model typically has to be differentiable to calculate the gradient.
Moreover, the value that converges by gradient-based methods depends on the initial value, and complex neural networks are prone to local minima.
Although the robot's range of motion is predetermined, gradient-based methods can cause it to exceed its range of motion unless some effort is made to devise a cost function.
Furthermore, in the task, the inverse model to optimize the joint angle $J^{\ast}_{t+1}$ from $I_{target}$ is expected to have numerous solutions.
However, neural networks do not acquire a perfect forward model.
Due to the nature of gradient-based methods, the solution is calculated where the error is smaller due to the imperfect forward model; only solutions that converge at a few possible solutions can be optimized.

The challenges can be summarized by the following four points:
1) Due to uncertainty, the deterministic approach in a task with dynamic motion causes a gap between the prediction and the robot motion.
2) The neural network model has to be differentiable to calculate the gradient and the gradient approach falls into the local minima.
3) It is time-consuming to apply arbitrary constraints with gradient-based optimization without affecting its performance or correctness.
4) There are numerous solutions for inverse kinematics (IK).

Our approach to these four challenges and the contributions of this paper are the followings.
1) We formulate the task of cable manipulation as a stochastic forward model. We then propose a method to address the uncertainty that maximizes the expected value, considering estimation errors.
2) The method introduces black-box optimization (BBO), non-differentiable optimization, to optimize $J^{\ast}_{t+1}$ without the need for gradients. 
Among BBO, we use the Tree-structured Parzen Estimator (TPE)~\cite{bergstra2011algorithms, bergstra2013making}, a type of Bayesian optimization.
3) TPE also easily incorporates arbitrary constraints on the range of solutions to optimize within the range of motion.
4) Since TPE is population-based, it can better detect multiple feasible configurations using the estimated inverse model.
Note that if the conditions from 2) to 4) of the challenges are satisfied, then anything other than a TPE can be used as a BBO.
\section{Method}
\label{sec:method}
Given a current joint angle $J_{t}$ and image $I_{t}$, and a target image $I_{target}$, our proposed method optimizes the target joint angle $J^{\ast}_{t+1}$ by considering uncertainty to realize the target image $I_{target}$ with high accuracy.
The proposed method consists of training and inference phases.
\subsection{Training}
\label{sec:Training}
The training phase is supervised learning in which the system can be trained to predict image $I_{t+1}$ when image $I_{t}$, joint angle $J_{t}$, and joint angle $J_{t+1}$ are given as inputs.
In other words, this is equivalent to learning the forward dynamics of the cable.
Considering this in a graphical model can be shown as Fig.~\ref{fig:setup}~(b).
The neural network's architecture based on this graphical model is shown in Fig.~\ref{fig:method}~(a).
The formulation of the training of this graphical model is described below.

n number of datasets $\mathbf{x} = \{ I_{t}^{i},\ I_{t+1}^{i},\ J_{t}^{i},\ J_{t+1}^{i}\}_{i=1}^{n}$ comprises the images $I_{t}$ \& $I_{t+1}$ and joint angles $J_{t}$ \& $J_{t+1}$ at step $t$ and $t+1$.
Let $P_{\theta}(x)$ estimate the true probability $P_{data}(x)$ of any $x$.
Then the maximum likelihood estimation of $\theta$ is defined by the following equation:
\begin{align}
        \         & \argmax_{\theta} \ \log{\prod}_{i}{P_{\theta}(I^{i}_{t+1}|I^{i}_{t}, J^{i}_{t}, J^{i}_{t+1})} \notag \\
        = \       & \argmax_{\theta} \ \sum_{i}{\log{P_{\theta}(I^{i}_{t+1}|I^{i}_{t}, J^{i}_{t}, J^{i}_{t+1})}} \notag \\
        \propto \ & \argmin_{\theta} \ \frac{1}{2n}\sum_{i}{(I^{i}_{t+1}-f_{\theta}(I^{i}_{t}, J^{i}_{t}, J^{i}_{t+1}))^{2}} \notag \\
        = \       & \argmin_{\theta} \ \frac{1}{2n}\sum_{i}{(I^{i}_{t+1}-I^{i}_{inferred})^{2}}
    \label{eq:training_eq}
\end{align}
where $f_{\theta}(\cdot)$ is function to predict $I_{inferred}$ from $I^{i}_{t}, J^{i}_{t}, J^{i}_{t+1}$.
\subsection{Inference}
\label{sec:Inference}
In the inference phase, the neural network model trained in the previous section is given a current image $I_{t}$ and joint angle $J_{t}$, and a target image $I_{target}$, and the joint angle $J^{\ast}_{t+1}$ is optimized so that the neural network predicts $I_{inferred}$ it is close to the target image $I_{target}$.
The graphical model in inference is the flow shown by the red arrow in Fig.~\ref{fig:setup}~(b).
A schematic is shown in Fig.~\ref{fig:method}~(c).
We will formulate this.
For readability, $I^{i}_t$, $J^{i}_{t}$, and $I^{i}_{t+1}$ are written as $I_{t}$, $J_{t}$, and $I_{t+1}$.
Given a current image $I_{t}$ and joint angle $J_{t}$, and a target image $I_{target}$, the goal is to optimize the joint angle $J^{\ast}_{t+1}$ to realize $I_{target}$, the joint distribution can be written as follows.
\begin{align}
              &\log{P(I_{t}, J_{t}, I_{t+1}, J^{\ast}_{t+1})} \notag \\
    \simeq    \    &\log{\lbrace
                            P_{\theta}(I_{t+1}|I_{t}, J_{t}, J^{\ast}_{t+1})
                            P_{\theta}(J^{\ast}_{t+1})
                            P_{\theta}(I_{t}, J_{t})
                        \rbrace}\notag \\
    =    \    &\log{\lbrace
                        P_{\theta}(I_{t+1}|I_{t}, J_{t}, J^{\ast}_{t+1})
                        P_{\theta}(J^{\ast}_{t+1})
                      \rbrace} + const.
\label{eq:infernce_eq1}
\end{align}
This leaves it vulnerable to uncertainty.
We maximize the likelihood expectation for this equation by considering the estimation error rather than the log-likelihood itself.
This study applies a stochastic approach, as described in \Cref{sec:introduction}.
The joint angle $J_{t+1}$ is reparameterized by $\mu+\sigma^2\epsilon$, where $\epsilon \sim {N(0, I)}$, $\mu$ is the average joint angle, and $\sigma^2$ is variance as a parameter for handling uncertainty.
\begin{align}
        \    &\log{\int{
                        P_{\theta}(I_{t+1}|I_{t}, J_{t}, \mu+\sigma^2\epsilon)
                        P_{\epsilon \sim {N(0, I)}} (\mu+\sigma^2\epsilon)
                      }}d\epsilon\notag \\
    \geq    \    &\int{P_{\epsilon \sim {N(0, I)}} (\mu+\sigma^2\epsilon)
                    \log{
                        P_{\theta}(I_{t+1}|I_{t}, J_{t}, \mu+\sigma^2\epsilon)
                      }}d\epsilon\notag \\
                & (\because \ Jensen's \ inequality)\notag \\
    =    \ &\mathbb{E}_{P_{\epsilon \sim {N(0, I)}}}
                        \lbrack
                            \log P_{\theta}(I_{t+1}|I_{t}, J_{t}, \mu+\sigma^2\epsilon)
                        \rbrack \notag \\
    \simeq \ &\frac{1}{2m}\sum_{i}^{m}{ (I_{t+1}-f_{\theta}(I_{t}, J_{t}, \mu+\sigma^2\epsilon_{i}))}^{2} + const.\notag \\
    := \ & g_{1}(\mu, \sigma)
    \label{eq:infernce_eq2}
\end{align}
where $m$ is the number of sampling.
Although the goal is to minimize $g_1$, $\sigma^2$ is introduced to address the uncertainty.
The smaller this value is, the smaller the effect of uncertainty will be treated as negligible.
Therefore, the approach becomes deterministic when the variance $\sigma^2$ is set to 0.
Under the given conditions of $I_{t}$ and $J_{t}$, the smaller the change in $I_{inferred}$ is for a significant change in $J^{\ast}_{t+1}$, the more robust it is against uncertainty.
Therefore, we will compute a multi-objective optimization of the following two functions.
\begin{equation}
        \argmin_{\mu \in [-90, \ 90]} g_{1}(\mu, \sigma), \ where \ \argmax_{\sigma \in [0.0, \ 1.0]} \sigma \\
    \label{eq:multi-objective black-box optimization}
\end{equation}
Note that the joint angles and variances range can be freely changed depending on the task and the experiment.
Multi-objective black-box optimization (MOBBO) is used for this multi-objective optimization.
Since it is an optimization of two cost functions, it will have a Pareto front.
\section{Experimental Setup}
\label{sec:experimental setup}
The task setting of this research is to manipulate a cable with a robotic arm.
Given the current joint angle of the robot $J_{t}$ and image of the cable $I_{t}$, and a target image of the cable $I_{target}$, the task is to optimize the joint angle $J^{\ast}_{t+1}$ to realize the target image $I_{target}$.
\subsection{Robot Setup for Data Collection System}
\label{sec:Robot Setup for Data Collections System}
Our robotic system, shown in Fig.~\ref{fig:setup}, consists of a self-designed 3-DoF robotic arm with a cable attached to the robot arm's end-effector.
Three Dynamixel XM430-W350-R are used for the robot's actuator.
Furthermore, we use an Intel RealSense Depth Camera D415 to overlook the workspace of the robot arm and cable and use them to capture RGB images.
Note that no depth information is retrieved in this study.
The robot and Realsense are connected to a MacBook Pro PC 2021 model running macOS Monterey.
The camera is mounted at the height of 650~mm from the workplace.
The workspace size for the cable manipulation is 785~mm${\times}$515~mm.
The length of the robot is 220~mm, and the length of the cable is 250~mm.
\subsection{Data Collection Process}
\label{sec:Data Collection Process}
We built a data collection system to collect data on cable manipulation.
The data collection process is as follows.
First, an image of the cable in the workspace is captured using Realsense installed on the ceiling.
At the same time, the current joint angles of the robot are recorded.
Next, the target joint angle for the next step is randomly selected in the range of $[-90 \ 90]$.
Note that the selected process is repeated until the end-effector position is within the workspace.
The robot's end-effector position is calculated based on the selected target joint angles.
Then, the robot executes the motion using the selected target joint angles.
Since the robot moves quickly, the cable moves with inertia.
However, since the recording image and the robot's joint angle are performed at 1~$Hz$, the image and joint angles are recorded after the robot and cable motion completely stop.
The discrepancy between the selected target joint and the actual motion is negligible.

This data collection process was performed for 12001 seconds, and 12001 images and corresponding joint angles of the robot were collected.
From this data, 12000 datasets were created from which the current joint angle $J_{t}$ and image $I_{t}$, the following joint angle $J_{t+1}$ and image $I_{t+1}$, were paired.
The joint angles were normalized to be $[0 \ 1]$.
The images were binarized by setting a threshold so that only the cable was extracted from the image.
Then, because the cable was too thin to train the neural network with this image well, image processing was performed to make the cable thicker using the cv2.dilate() function with the kernel size (3, 3).
The original RGB image at time $t$ and $t+1$, the binarized image, and the image processed to make it thicker are shown in Fig.~\ref{fig:setup}.
Of the 12,000 samples collected, 10,000 were used for training, and 2,000 were used for validation.

For inference, the robot's current joint angle $J_{t}$, the cable's image $I_{t}$, and the target cable's image $I_{target}$ are given to the trained neural network, and the joint angle $J^{\ast}_{t+1}$ to realize the target image $I_{target}$ is optimized.
This optimized joint angle $J^{\ast}_{t+1}$ is given as the command value of the robot's target joint angle, and the robot moves to the joint angle.
\subsection{Deep Learning \& Training \& Inference}
\label{sec:Deep Learning}
Our neural network model comprises CNNs and FCs layers, and details of the neural network architecture are shown in Fig.~\ref{fig:method}.
We used PyTorch as a deep learning library and Optuna~\cite{optuna_2019} as a BBO library for implementation.
We trained the neural networks and conducted inference on a machine equipped with an Intel Xeon Gold 6524 CPU @ 3.10~$GHz$ and Tesla V100 SXM2 with 16GB.
Training for the model is about 5 minutes.

\section{Experiment Results}
\label{sec:results}
In this experiment, we evaluate the following four items.
1) Comparison of optimization approaches between gradient-based approaches and BBO (\Cref{sec: Comparison between Gradient Approach and Black-Box Optimization}).
2) Comparison of gradient-based methods and BBO when the robot has numerous solutions with redundant degrees of freedom (\Cref{sec: Multiple plots of Gradient Approach and Black-Box Optimization}).
3) Evaluate when using the proposed method, MOBBO (\Cref{sec: Multi-Objective Black-Box Optimization}).
4) Evaluate the performance of the proposed method when the cable manipulation is executed by the robot (\Cref{sec: Robot Experiment}).
\begin{figure}[tb]
    \centering
    \includegraphics[width=0.80\columnwidth]{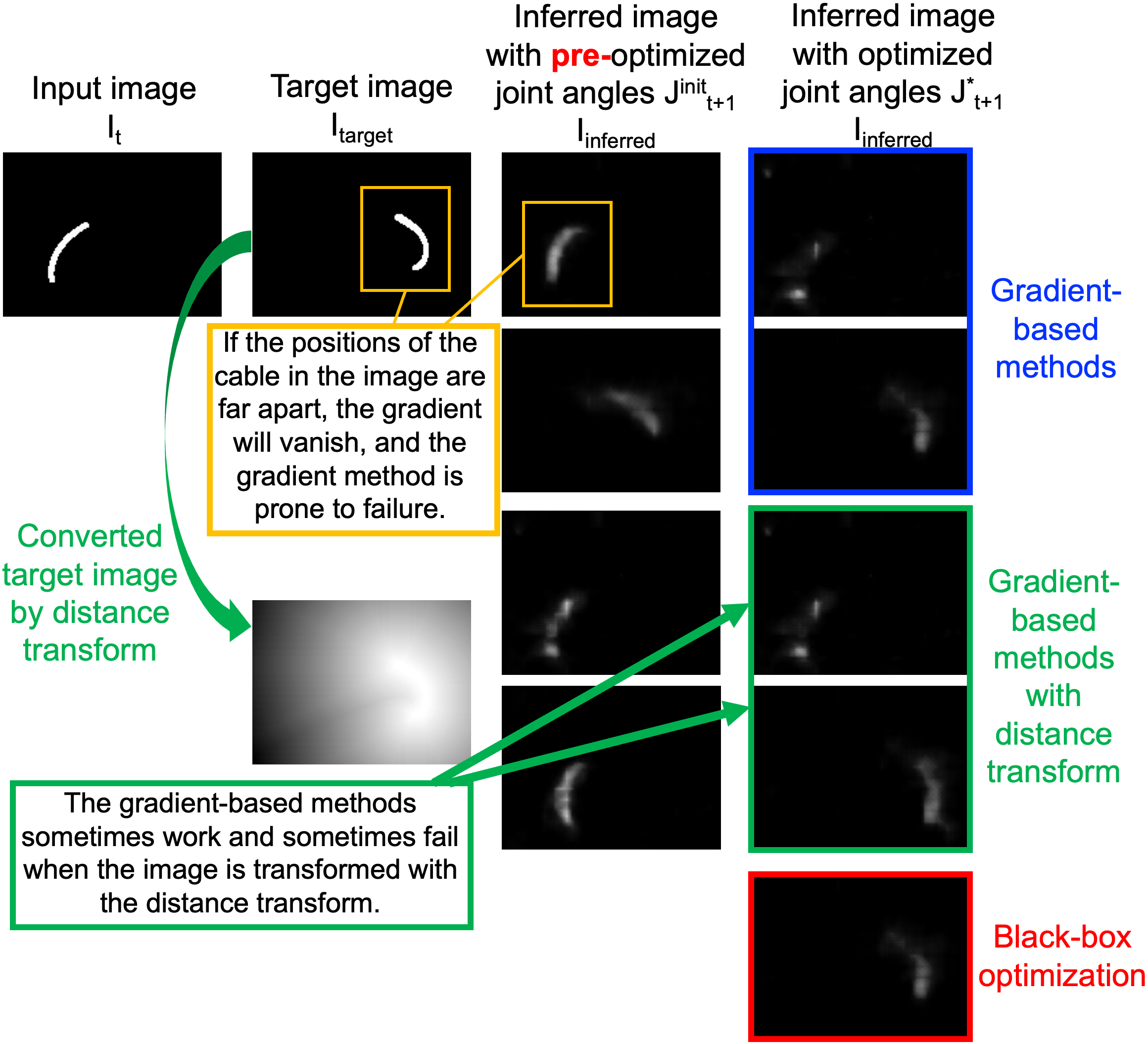}
    \caption{
            Comparison of inferred image $I_{inferred}$ with optimized joint angle $J^{\ast}_{t+1}$ using gradient-based methods, distance transform, and BBO.
            Note that because of the uncertainty in the cable's behavior, inferred image $I_{inferred}$ of the cable will be blurry.
            }
    \label{fig:image_distance}
\end{figure}
\subsection{Comparison between Gradient-based Methods and BBO}
\label{sec: Comparison between Gradient Approach and Black-Box Optimization}
This section compares the methods using gradient and BBO.
For the optimization using gradient-based methods, a randomly set initial value was given as the joint angle $J^{init}_{t+1}$ in the manner described in \Cref{sec:introduction}.
For the method using BBO, the variance $\sigma^2$ of the proposed method was set to 0 in eq.~\eqref{eq:multi-objective black-box optimization}, which means treated deterministically rather than stochastically.
Therefore, the cost functions of gradient-based methods and BBO are the same design, but only the optimization methods differ, and it can evaluate the differences between the two optimization methods.

Examples of pairs of inferred images $I_{inferred}$ using gradient-based methods, the target image $I_{target}$, and images inferred with randomly set initial values $J^{init}_{t+1}$ are shown in Fig.~\ref{fig:image_distance}.
As shown in Fig.~\ref{fig:image_distance} on the \nth{1} row, there are entirely different between the target image $I_{target}$ and the inferred image $I_{inferred}$ using optimized joint angle $J^{\ast}_{t+1}$ to minimize the difference between target image $I_{target}$ and inferred image $I_{inferred}$.
From the figures on the \nth{2} row, it can also be seen that it depends on the initial values $J^{init}_{t+1}$.
The reasons for the failure of gradient-based methods can be explained as follows.
When using MSE to calculate from the target image $I_{target}$ and the inferred image $I_{inferred}$ using $J^{init}_{t+1}$ at the beginning of optimization of the joint angle or in the process of optimization, if the pixels where the cable exists do not match, the error value is treated as the same whether the cable is misaligned by 1 pixel or 100 pixels.
If the position of the cable in the inferred image $I_{inferred}$ and the target image $I_{target}$ are far apart, the slight change in joint angle will not change MSE values, and the gradient will vanish (Compare the \nth{1} and \nth{2} raw in Fig.~\ref{fig:image_distance}).

To address this challenge, \cite{kawaharazuka2019dynamic} transformed the target image $I_{target}$ into a distance image to avoid gradient vanishing instead of a binary 0 and 1 image with the representation of cable present or absent.
The results of gradient-based methods on an image converted to a distance image by the cv2.distanceTransform() function are shown in Fig.~\ref{fig:image_distance} on the \nth{3}-\nth{4} row. 
There are cases where it works and fails, and it can be seen that it depends on the initial values of the joint angles $J^{init}_{t+1}$.
It may be that the latent expression is too complex and is prone to local minima.

Next, the results for the case using BBO are shown in Fig.~\ref{fig:image_distance} on the bottom row.
Even in cases where gradient-based methods failed, the method using BBO successfully inferred an image $I_{inferred}$ that was close to the target image $I_{target}$.
Since the initial TPE (BBO) sampling stage explores a broader parameter space, the possibility of overlap between the image inferred from the sampled joint angles and the pixels in the target image $I_{target}$ increases.
As a result, even when gradient-based methods fail, BBO can optimize joint angles that produce an inferred image $I_{inferred}$ closer to the target image $I_{target}$.
\begin{figure}[tb]
    \centering
    \includegraphics[width=0.85\columnwidth]{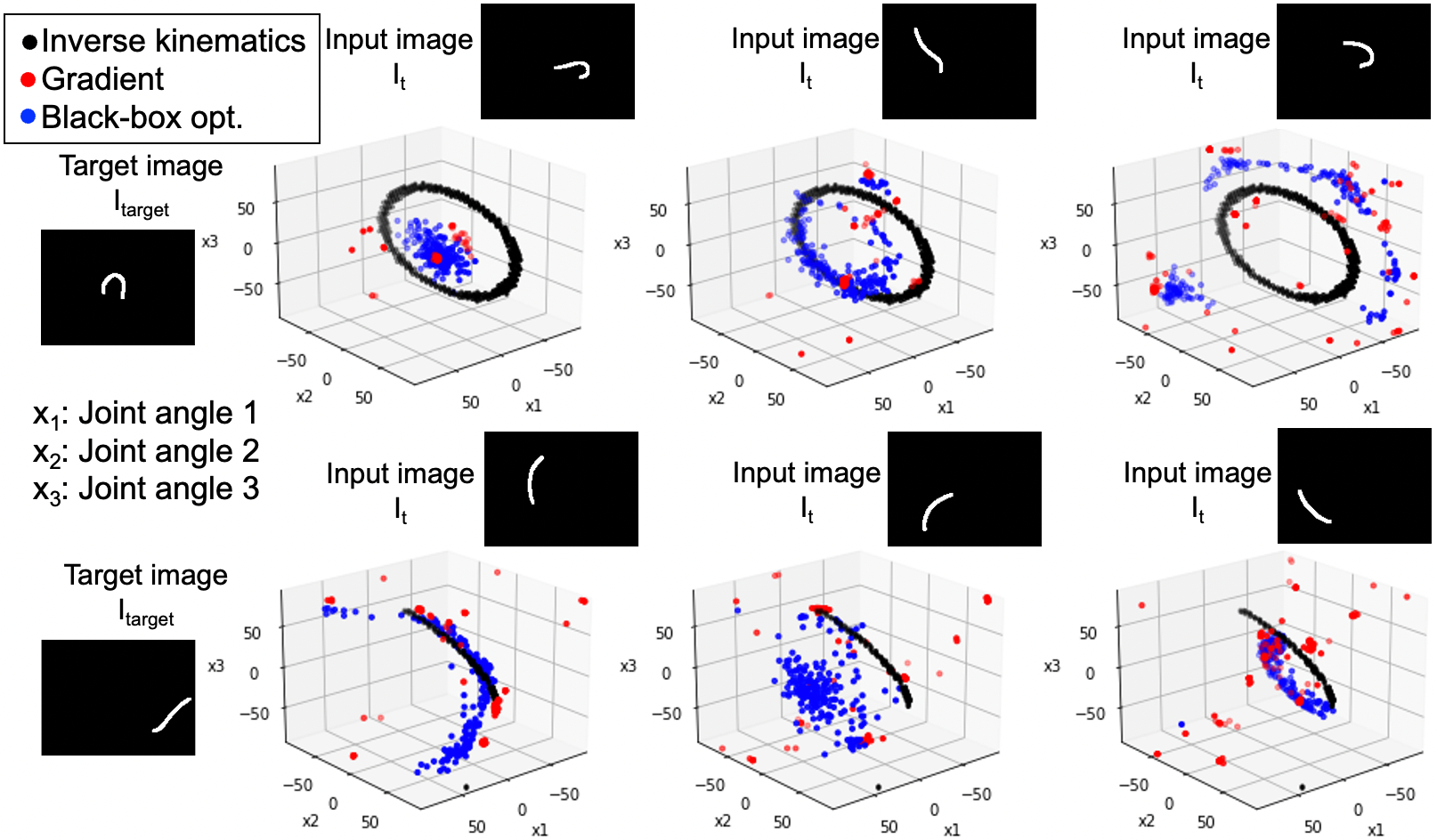}
    \caption{
            Plots of approximate solution of IK and optimized joint angles $J^{\ast}_{t+1}$ using gradient method and BBO when the target images $I_{target}$ are the same, but image $I_{t}$ and joint $J_{t}$ are different.
            }
    \label{fig:multiple_plots}
\end{figure}

\subsection{Multiple plots of Gradient-Based methods and BBO}
\label{sec: Multiple plots of Gradient Approach and Black-Box Optimization}
In the task of the experiment in this paper, the robotic arm with 3-DoF executes the motion in the two-dimensional plane.
Therefore, there are numerous solutions for solving inverse kinematics (IK) given the end-effector coordinates (x, y).
We examine how gradient-based methods and BBO behave for such a task.

For gradient-based methods, optimization of joint angle with randomly given initial values $J^{init}_{t+1}$ was performed with 200 identical pairs as a current joint angle $J_{t}$, image $I_{t}$, and target image $I_{target}$.
Even in BBO, the sampling values of the joint angle $J^{init}_{t+1}$ to be optimized start from random values.
Approximate solutions for IK were also calculated from the end-effector coordinates calculated by forward kinematics using the joint angles when the target image $I_{target}$ data was collected.
To calculate the approximate solution for IK, the joint angles within $\pm\delta$ for the end-effector coordinates were searched by grid search.
The optimized joint angles $J^{\ast}_{t+1}$ for several pairs of different joint angles $J_{t}$, images $I_{t}$, and target images $I_{target}$ are plotted in Fig.~\ref{fig:multiple_plots} using the same target image $I_{target}$ but differ in joint angles $J_{t}$ and images $I_{t}$.
The black plots are approximate solutions for IK, the blue plots are those calculated by BBO, and the red plots are those calculated using gradient-based methods.

In this experimental setup, the solution of IK is basically an elliptical trajectory.
However, due to the limitation of each joint motion range, $[-90 \ 90]$, only a portion of the elliptical trajectory may be plotted.
Other than that, the positions of the end-effector coordinates may be points due to the robot's posture.
It can be seen that the optimized joint angles　$J^{\ast}_{t+1}$ are distributed as a region when BBO is used (Fig.~\ref{fig:multiple_plots}).
IK uses only the robot's end-effector coordinates at step $t$, so the solution is different from the BBO method, which calculates the optimized joint angle $J^{\ast}_{t+1}$ from a current image $I_{t}$, joint angle $J_{t}$, and target image $I_{target}$, but the BBO method has numerous solutions.
In addition, when the target image $I_{target}$ is the same, but the joint angles $J_{t}$ and images $I_{t}$ differ, optimized joint angles $J^{\ast}_{t+1}$ are changed when BBO is used.
This indicates that the neural network acquires a forward model of cable manipulation that considers the dynamics and can be calculated as an inverse model using BBO.
On the other hand, with gradient-based methods, the optimized joint angles $J^{\ast}_{t+1}$ converge within a narrow range at several locations (The red plots in Fig.~\ref{fig:multiple_plots}).
Theoretically, numerous optimized joint angles $J^{\ast}_{t+1}$ are plotted as trajectories as in the BBO.
The neural network does not acquire a perfect forward model; thus, slight differences exist in each value, even if they can theoretically be the same.
Due to the characteristics of gradient-based methods, the solution is calculated where the error is smaller due to the imperfect forward model, so it converges to a point as shown in Fig.~\ref{fig:multiple_plots}).
When the optimization starts, the points that converge depend on the initial values of joint angle $J^{init}_{t+1}$.
Some fall into local minima, resulting in entirely different between target images $I_{target}$ and images $I_{inferred}$ inferred using the optimized joint angles $J^{\ast}_{t+1}$ by gradient-based methods.
Note that when numerous solutions exist, using the ensemble method to reduce the bias and variance of the neural network and calculating the average of the optimal solution output by each model does not necessarily give good results.
It can be said that using a neural network does not always mean that gradient-based methods are a good choice.
The proposed method is expected to decrease the model's bias and variance and will be validated in the \Cref{sec: Robot Experiment}.
\begin{figure}[tb]
    \centering
    \includegraphics[width=0.80\columnwidth]{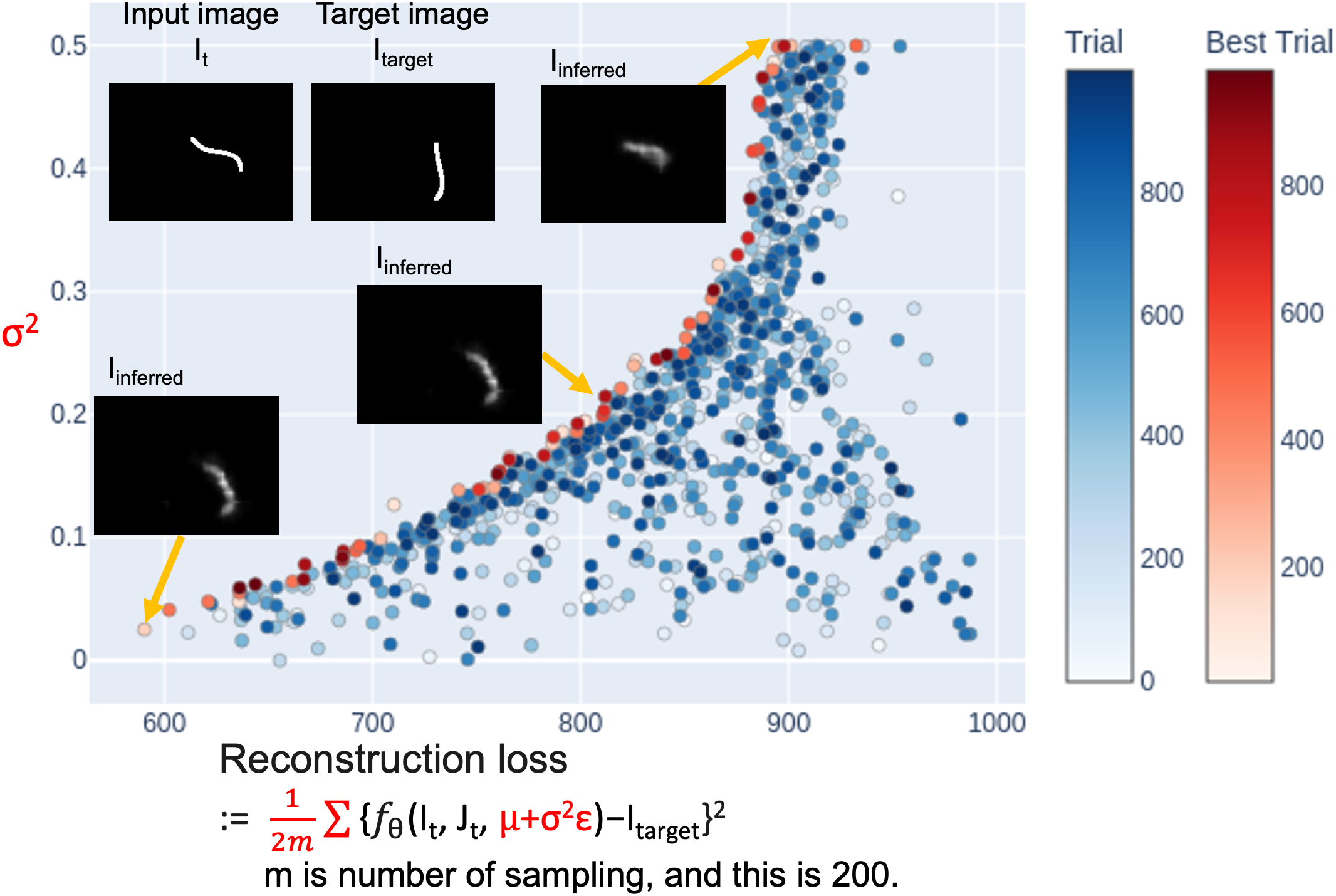}
    \caption{
            Plots of Pareto solution with multi-objective black-box optimization
            }
    \label{fig:paretp-front}
\end{figure}

\subsection{Multi-Objective Black-Box Optimization}
\label{sec: Multi-Objective Black-Box Optimization}
In the proposed method, multi-objective black-box optimization (MOBBO) is performed with joint angle $J_{t+1}$ and variance $\sigma^2$ as optimization parameters, as shown in eq.~\eqref{eq:multi-objective black-box optimization}.
A plot of the Pareto solution of the MOBBO for a given current image $I_{t}$, joint angle $J_{t}$, and the target image $I_{target}$, is shown in Fig.~\ref{fig:paretp-front}.
In this MOBBO, the number of trials is 1000, and the range of each joint angle is searched in the integer range as $[-80 \ 80]$ and the range of variance as $[0 \ 0.5]$ in eq.~\eqref{eq:multi-objective black-box optimization}.
The number of sampling $m$ is 200 in eq.~\eqref{eq:infernce_eq2}.
The Pareto front is shown in the red plot.
The $I_{inferred}$ shown in Fig.~\ref{fig:paretp-front} is calculated with reparameterized joint angle $\mu+\sigma^2\epsilon \ (=J_{t+1})$ of $\mu$.
As the variance $\sigma^2$ increases, the reconstruction loss in eq.~\eqref{eq:infernce_eq2} increases because the difference from the target image $I_{target}$ is more significant in the calculation.
However, when the variance $\sigma^2$ is 0, if there is uncertainty in the trained neural network and robot environment, there would be a probability that the target image $I_{target}$
would not be close to the image $I_{robot}$ when the robot moves using optimized joint angle $J^{\ast}_{t+1}$.
Therefore, selecting an appropriate variance $\sigma^2$ for the model from the Pareto front becomes necessary.
In this study, optimization is performed in the range of variance $[0.03 \ 0.05]$, and the robot is tested using the optimized joint angle $J^{\ast}_{t+1}$ with the lowest reconstruction loss (See \Cref{sec: Robot Experiment}).

\begin{table}[t]
    \centering
    \caption{
            The cost is calculated using EMD among the target image $I_{target}$, the inferred image $I_{inferred}$, and the images obtained by the robot's motion $I_{robot}$. And the percentage of the EMD cost below the thresholds.
            }
    \label{tab: result_robot_experiment}
    \begingroup
    \scalefont{0.80}
        \begin{tabular}{c|c||c | c c }
        \hline
        EMD b/w & Methods & EMD \ $mean\pm std$ & w/n 125 & w/n 175\\
        \hline\hline
        Target- &
        Gradient
                          & $2103.1\pm2842.9$ & $20.0\pm4.6$ & $30.0\pm5.3$\\
        Inferred &
        BBO
                          & $528.8\pm1557.5$ & $38.1\pm5.6$ & $52.7\pm5.7$\\
        &Ours
                          & $\bf{452.5}\pm1533.9$ & $\bf{46.7}\pm5.8$ & $\bf{58.7}\pm5.7$ \\
        \hline

        Target- &
        Gradient
                          & $2590.7\pm3216.4$ & $17.3\pm4.3$ & $20.6\pm4.6$ \\
        Robot &
        BBO
                          & $722.3\pm1786.7$ & $28.0\pm5.2$ & $36.7\pm5.5$ \\
        &Ours
                          & $\bf{569.5}\pm1531.6$ & $\bf{30.0}\pm5.3$ & $\bf{42.0}\pm5.7$ \\
        \hline

        Robot- &
        Gradient
                          & $ 429.7\pm 445.1$ & $26.0\pm5.1$ & $40.7\pm5.7$ \\
        Inferred &
        BBO
                          & $334.8\pm 402.3$ & $43.4\pm5.7$ & $54.0\pm5.8$ \\
        &Ours
                          & $\bf{241.1}\pm 343.1$ & $\bf{53.3}\pm5.8$ & $\bf{64.7}\pm5.5$ \\
        \hline
        \end{tabular}
    \endgroup
\end{table} 

\begin{figure}[tb]
    \centering
    \includegraphics[width=0.70\columnwidth]{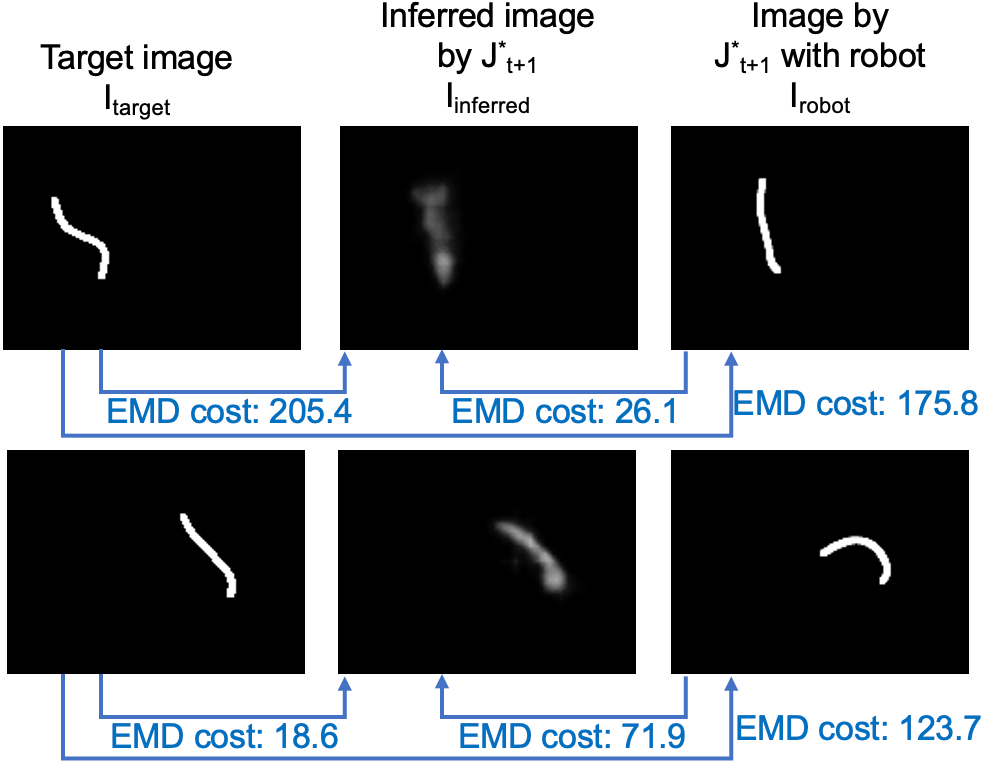}
    \caption{Example of cost value by EMD}
    \label{fig:emd}
\end{figure}
\subsection{Robot Experiment}
\label{sec: Robot Experiment}
This section shows the evaluation results of cable manipulation with the robot motion.
The following three methods are evaluated: gradient-based methods, BBO, which is the proposed method with variance $\sigma^2$ set to 0 in eq.~\eqref{eq:multi-objective black-box optimization}, and the proposed method.
$\sigma^2=0$ in eq.~\eqref{eq:multi-objective black-box optimization} means that the method is treated deterministically rather than stochastically.

To prepare the target image $I_{target}$, the joint angle $J_{t}$, and the image of the cable $I_{t}$ with diversity in the state of the cable, we let the robot perform 6 steps.
The third step was set as the target image $I_{target}$, the joint angle of the robot after 6 steps were set as $J_{t}$, and the image of the cable was set as $I_{t}$.
Then, optimized joint angle $J^{\ast}_{t+1}$ is calculated.

If MSE is used as an evaluation method, a 1-pixel and 100-pixel shift are considered the same error, making it difficult to evaluate the result appropriately.
Therefore, the optimal transport distance, Earth Mover's Distance (EMD), is used.
EMD is also known as the Wasserstein metric.
Given image A and image B, it is the cost of minimizing the product of the pixel shift distance and the shift amount to make image A resemble image B.
Note that since EMD is computationally expensive, it could not be used as a substitute for the MSE, which is used for training errors, inference, and MOBBO costs in this study.

This EMD is used to compare three images.
(1) the target image $I_{target}$ and the inferred image $I_{inferred}$, (2) target image $I_{target}$ and the image of the cable obtained when the robot is moved $I_{robot}$ using the optimized joint angles $J^{\ast}_{t+1}$, and (3) $I_{robot}$ and $I_{inferred}$.
In (1), it can be evaluated whether realizing the target image $I_{target}$ is difficult with a one-step motion or whether the optimal joint angle using gradient-based methods falls into the local minima.
(2) evaluates how close the target image $I_{target}$　is to the image $I_{robot}$ captured by the robot's motion with the optimized joint angles $J^{\ast}_{t+1}$, which allows the method's effectiveness to be assessed.
In (3), the accuracy of the neural network's predictions can be evaluated by assessing how close the inferred image　$I_{inferred}$ is to the image $I_{robot}$ obtained from the actual robot's motion.

The results of the evaluation with these values are shown in Table~\ref{tab: result_robot_experiment}.
For each method, 150 trials of motion were performed to evaluate the results.
The mean and variance of the cost of EMD are shown.
When the images $I_{robot}$ captured when the robot executed the motion deviated from the target image $I_{target}$ significantly, the value of the cost of EMD sometimes became huge, resulting in a large mean and variance.
Therefore, we also calculated the percentages of EMD cost values below 125 and 175, respectively.
We calculated the mean and standard deviation of success rates by bootstrapping.
The EMD cost values less than 125 to 175 are similar to the target image $I_{target}$ and the images $I_{robot}$ taken by the robot as it moves the cable (Fig.~\ref{fig:emd}).
Although the EMD does not take cable topology into account, the results of this experiment suggest that if the EMD cost value is around 125-175, the cable misalignment is slight, and topology is less of an issue.

When gradient-based methods are used, the average cost of EMD between Target-Inferred is much larger than the other two.
This indicates that gradient-based methods fall into the local minima, as mentioned in \Cref{sec: Comparison between Gradient Approach and Black-Box Optimization}.
Next, we compare the proposed method with the BBO (deterministic approach) case between Robot-Inferred.
The proposed method has a smaller value of the average EMD cost and larger values of w/n 125 and w/n 175 compared to BBO.
In other words, if the optimal joint angles $J^{\ast}_{t+1}$ using the deterministic approach (BBO) are selected and the robot moves, the trained neural network has uncertainties, leading to misalignments.
In contrast, the proposed method considers these uncertainties to increase the probability of achieving an image $I_{robot}$ close to the target image $I_{target}$.
As a result, the proposed method performs better than the other methods when the robot performs motions.
This shows the effectiveness of the proposed method.
\section{Conclusion}
\label{sec:conclusion}
This paper formulated dynamic cable manipulation as a stochastic forward model to realize the target image.
We proposed a method to handle uncertainties that maximizes the expectation and considers estimation errors.
This method avoids the risk of falling into local solutions by applying BBO.
Moreover, by incorporating the BBO, namely TPE, as a constraint, we could obtain a solution within the range of motion.
Furthermore, since TPE is population-based, we were able to confirm that it can handle numerous solutions.
We used EMD to evaluate the similarity of the cable's image $I_{robot}$ to the target image $I_{target}$ when the robot moves.
The results confirmed the improved accuracy of the proposed method compared to conventional methods, such as the gradient-based methods and the deterministic approach that does not consider uncertainty.
\section*{ACKNOWLEDGMENT}\small
The authors would like to thank Dr. Shin-ichi Maeda, Avinash Ummadisingu, and Dr. Naoki Fukaya for the many discussions about this research.
This work was supported by JST [Moonshot R\&D][Grant Number JPMJMS2033].
\bibliographystyle{IEEEtran} 
\bibliography{IEEEabrv,bibliography}
\end{document}